\documentclass[runningheads]{llncs}

\usepackage[mobile]{eccv}

\usepackage{eccvabbrv}
\usepackage{xspace}
\usepackage{float}
\usepackage{graphicx}
\usepackage{wrapfig}
\usepackage[export]{adjustbox}                  %
\usepackage{booktabs}
\usepackage{multirow}
\usepackage[pagebackref,breaklinks,colorlinks,citecolor=eccvblue]{hyperref}
\usepackage{soul}
\usepackage[accsupp]{axessibility}  %

\newcommand{\methodname}{Multi-Object Decoder\xspace}
\newcommand{\benchmark}{MessyKitchens\xspace}

\usepackage{orcidlink}

\begin{document}

\title{MessyKitchens: Contact-rich object-level\\ 3D scene reconstruction}
\titlerunning{MessyKitchens}

\renewcommand{\thefootnote}{\fnsymbol{footnote}}

\author{
Junaid Ahmed Ansari\inst{1}\textsuperscript{*} \and
Ran Ding\inst{1}\textsuperscript{*} \and
Fabio Pizzati\inst{1} \and
Ivan Laptev\inst{1}
}

\authorrunning{J.~Ansari and R.~Ding et al.}

\institute{
Mohamed bin Zayed University of Artificial Intelligence, Abu Dhabi, UAE\\
\email{\{junaid.ansari,ran.ding,fabio.pizzati,ivan.laptev\}@mbzuai.ac.ae}\\
\centerline{\textsuperscript{*}Equal contribution}
}

\maketitle

\begin{abstract}    
Monocular 3D scene reconstruction has recently seen significant progress. Powered by the modern neural architectures and large-scale data, recent methods achieve high performance in depth estimation from a single image. Meanwhile, reconstructing and decomposing common scenes into individual 3D objects remains a hard challenge due to the large variety of objects, frequent occlusions and complex object relations.
Notably, beyond shape and pose estimation of individual objects, applications in robotics and animation require physically-plausible scene reconstruction where objects obey physical principles of non-penetration and realistic contacts.
In this work we advance object-level scene reconstruction along two directions. 
First, we introduce \mbox{\em MessyKitchens}, a new dataset with real-world scenes featuring cluttered environments and providing high-fidelity object-level ground truth in terms of 3D object shapes, poses and accurate object contacts.
Second, we build on the recent SAM 3D approach for single-object reconstruction and extend it with Multi-Object Decoder (MOD) for joint object-level scene reconstruction.
To validate our contributions, we demonstrate MessyKitchens to significantly improve previous datasets in registration accuracy and inter-object penetration. We also compare our multi-object reconstruction approach on three datasets and demonstrate consistent and significant improvements of MOD over the state of the art. Our new benchmark, code and pre-trained models are publicly available on our project website: \url{https://messykitchens.github.io/}. 

\end{abstract}

\begin{figure}[t]
  \centerline{
     \includegraphics[width=1.0\textwidth,trim={0 3.3cm 3cm 3.3cm},clip]{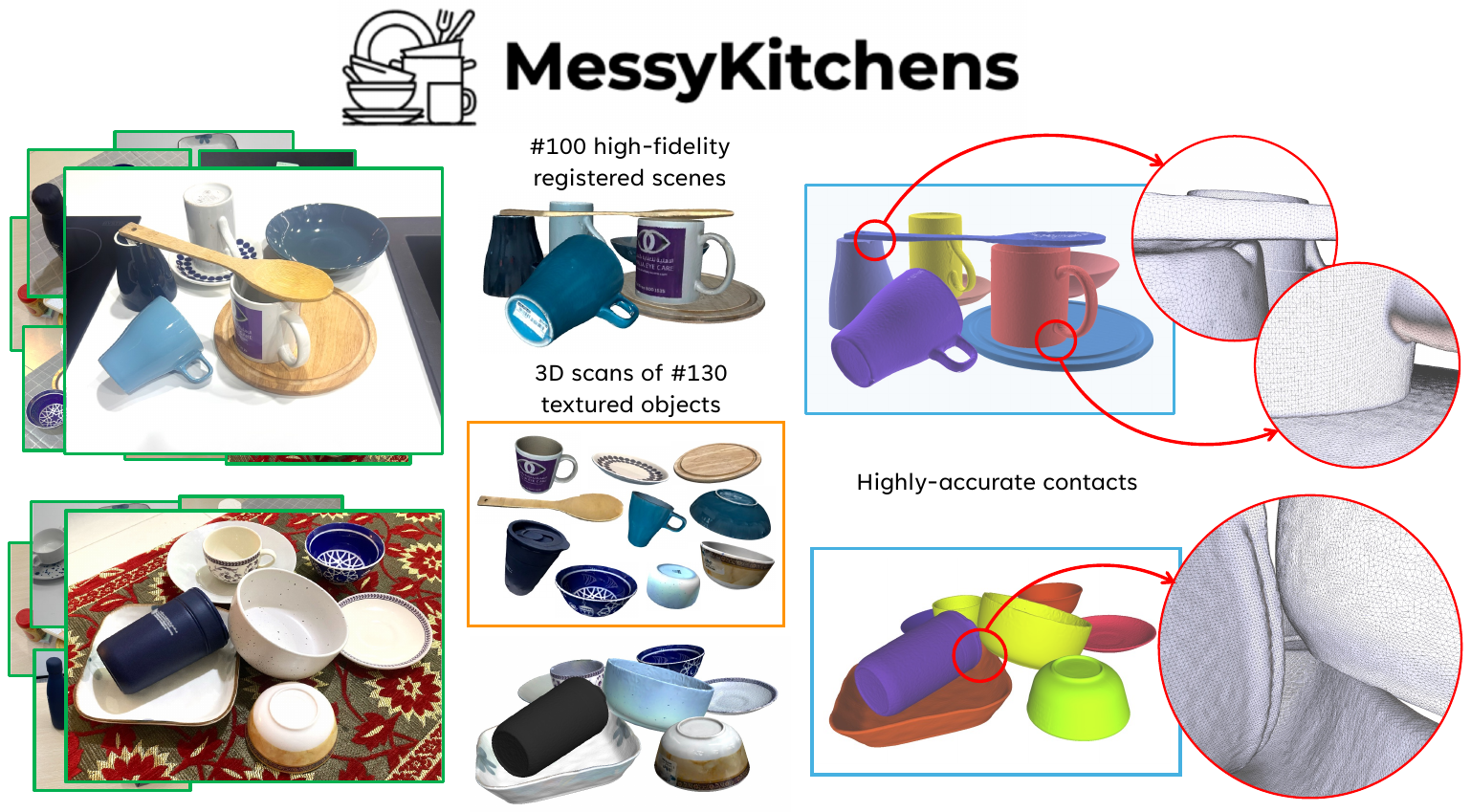}
    }
    \vspace{-.3cm}
    \caption{\textbf{MessyKitchens benchmark. } Images of real scenes and corresponding high-fidelity object-level 3D scenes reconstructions composed of accurate object scans.}
  \label{fig:teaser}
 \end{figure}

\section{Introduction}
Accurate 3D scene reconstruction plays a pivotal role in many applications in digital arts and content creation, industrial inspection, surgery, heritage preservation, navigation, as well as robot learning and simulation.
While some tasks, e.g., navigation~\cite{robnav1, robnav2, robnav3, robnav-4}, may only require free space estimation and scene-level surface reconstruction, other tasks, e.g., robotic manipulation~\cite{graspnet-1b, manip-1, manip-2, manip-3, manip-4, manip-5, manip-6}, motion planning~\cite{zhou2021weldingpath, do2023_geometry_aware_cpp, tiboni2025maskplanner, chen2026_3dcovdiffusion}, and animation~\cite{pumarola2021d, wu2025animateanymesh, blender}, often rely on detailed 3D representations of individual objects.
Object-level scene reconstruction is a difficult task challenged by the large variety of object shapes and frequent occlusions.
Moreover, beyond object shape and pose estimation, animation and simulation tasks require physically-plausible scene reconstruction with correct estimation of object contacts, deformations, and other parameters.

\noindent Recent methods for 3D scene reconstruction have evolved from classic geometry-based formulations to learning-based approaches. The latter methods rely on learned inductive biases and enable accurate shape predictions from a single image. In particular, several recent methods such as DepthAnything~\cite{yang2024depth}, VGGT~\cite{wang2025vggt} and Gen3C~\cite{ren2025gen3c} significantly advance results for monocular depth estimation.
In comparison, object-level scene reconstruction has received relatively less attention. Among existing methods, MIDI~\cite{huang2025midi} and PartCrafter~\cite{lin2025partcrafter} present impressive results, but focus on synthetic scenes, while the recent SAM 3D~\cite{chen2025sam} enables the estimation of the shape and pose of single objects in real images.
Besides new methods, the progress in object-level scene reconstruction also requires realistic and high-fidelity benchmarks for training and evaluation. While several benchmarks exist~\cite{graspclutter6d,graspnet-1b,housecat6d}, their 3D ground truth often suffers from the limited registration accuracy and inter-object penetrations.  

To address these limitations, we advance object-level scene reconstruction on two fronts. 
{\em First}, we introduce {\em MessyKitchens}, a new dataset with 100 real-world scenes featuring cluttered environments along with high-fidelity object-level 3D ground truth. As shown in Figure~\ref{fig:teaser}, our dataset contains contact-rich scenes and 130 varying kitchen objects that have been scanned and registered with high precision. Moreover, MessyKitchens provides accurate object contacts, and hence enables evaluation of geometrically-precise and  physically-realistic scene reconstruction. In addition, we also provide a large-scale synthetic training set MessyKitchens-synthetic composed of 1.8k contact-rich scenes and 10.8k rendered images.
{\em Second}, alongside the benchmark, we propose a method for joint object-level scene reconstruction. Our approach builds upon the recent SAM 3D framework for single-object reconstruction and extends it with a Multi-Object Decoder (MOD) that jointly predicts the geometry and poses of multiple objects in a scene. By reconstructing objects simultaneously, MOD captures contextual relationships and enforces more physically-plausible configurations.

In summary, we propose the following contributions:
\begin{itemize}
\item We introduce MessyKitchens, a new benchmark with cluttered real scenes and high-fidelity 3D object-level ground truth, including accurate object shapes, poses and contacts along with precise scene-level object registration. 
\item We propose Multi-Object Decoder (MOD), a method that extends SAM 3D to the joint modeling and reconstruction of multiple objects in a scene. %
\item Through extensive experiments we show MOD to outperform state-of-the-art object-level scene reconstruction in MessyKitchens, GraspNet-1B~\cite{graspnet-1b}, HouseCat6D~\cite{housecat6d}, and GraspClutter6D~\cite{graspclutter6d}.
We also demonstrate significantly improved 3D accuracy of the MessyKitchens benchmark compared to other recent datasets.
\end{itemize}

\section{Related Works}

\paragraph{Existing benchmarks for 3D scenes.}
Early datasets such as LINEMOD \cite{LINEMOD}, T-LESS\cite{T-LESS}, YCB-M\cite{YCB-M}, and YCB-Video \cite{YCB-V} established standardized evaluation protocols for 3D pose estimation and object-level reconstruction, significantly advancing the field. However, they were captured in controlled laboratory settings and typically contain a limited number of objects and scenes with restricted category diversity. Datasets such as MP6D \cite{MP6D} and T-LESS \cite{T-LESS} focus largely on industrial objects and lack representation of everyday kitchen categories, limiting their realism for domestic environments. Although synthetic datasets such as Falling Things \cite{tremblay2018falling} and ZeroGrasp-11B \cite{zerograsp-11b} scale up object count substantially, they remain purely simulation-based and do not reflect the challenges of real-world scene acquisition and annotation. More recent datasets, including GraspNet-1B \cite{graspnet-1b}, HouseCat6D \cite{housecat6d}, PhoCaL \cite{phocal}, Omni6DPose \cite{omni6dpose}, KITchen \cite{younes2024kitchen}, PACE \cite{you2023pace}, and GraspClutter6D \cite{graspclutter6d}, increase object diversity and incorporate kitchen and transparent objects. Nevertheless, they are often constrained by limited annotation accuracy, inaccurate contacts, insufficient scene complexity, or scalability (e.g., \cite{phocal, housecat6d} rely on robotic manipulator or motion capture systems for data collection). In contrast, MessyKitchens is collected using a portable 3D scanner enabling scene acquisition across diverse locations and achieves higher object-to-scene registration accuracy that naturally yields improved contacts, while remaining comparable in number and diversity of objects.

\paragraph{Object-centric 3D scene reconstruction.}
Object-centric 3D scene reconstruction has traditionally evolved from feed-forward and retrieval-based strategies. Feed-forward methods typically leverage encoder-decoder architectures to regress scene properties, such as geometry, instance labels, and poses, directly from images~\cite{nie2020total3d, zhang2021im3d, paschalidou2021atiss, dahnert2021panorecon, zhang2023uni3d, xie2019pix2vox, wang2018pixel2mesh, gkioxari2019mesh}, while retrieval-based approaches, such as DiffCAD~\cite{gao2024diffcad} and ACDC~\cite{dai2024automated}, align high-quality 3D assets from existing databases to input~\cite{gumeli2022roca, izadinia2017im2cad, kuo2020mask2cad, kuo2021patch2cad}. However, these methods are often limited by the scarcity of supervised 3D data and a heavy dependence on database diversity, which hinders their ability to generalize to complex out-of-distribution scenes~\cite{chang2015shapenet, deitke2023objaverse, deitke2024objaversexl}. Recently, compositional and modular generative paradigms~\cite{chen2024comboverse, han2024reparo, tang2024diffuscene, ardelean2025gen3dsr} have sought to model complex scenes utilizing large-scale perceptual and 3D priors~\cite{kirillov2023sam, ren2024groundedsam, rombach2022ldm, jun2023shape, eftekhar2021omnidata, zhang2024clay}. Frameworks such as MIDI~\cite{huang2025midi} and PartCrafter~\cite{lin2025partcrafter} achieve high generative fidelity through multi-instance diffusion and Diffusion Transformers (DiT)~\cite{peebles2023dit}. However, these approaches are primarily developed on synthetic datasets and often face challenges when generalizing to the complexities of real-world captures. In contrast, SAM 3D~\cite{chen2025sam} provides a highly scalable pipeline that delivers strong foundational results in diverse real-world scenes through robust data alignment. Yet, because SAM 3D typically processes objects as independent tokens, it lacks explicit, end-to-end reasoning regarding the spatial inter-dependencies between multiple objects. This independent treatment often leads to inaccuracies in the global spatial layout and imprecise relative object poses, posing a significant challenge for achieving spatially consistent 3D scene reconstructions in complex environments ~\cite{yao2025cast}. \looseness=-1

\section{The \benchmark Benchmark}

We introduce here our \benchmark benchmark. For it, our core aim is the evaluation of the accuracy of object-centered 3D scene reconstruction on challenging scenarios, using physically-curated ground truth data. We describe the process and the characteristics of the data in Section~\ref{sec:method-real}. In Section~\ref{sec:method-synth}, we also provide a \textit{synthetic} set, coined \benchmark-synthetic, to enable training on similar scenarios. Finally, in Section~\ref{sec:method-mod}, we propose \methodname as a new simple baseline for 3D scene reconstruction.

\subsection{Real data}\label{sec:method-real}
\paragraph{Data acquisition.} We now describe our data acquisition strategy. In total, we collect 100 real scenes, each composed of a variable number of kitchenware objects that depend on the difficulty level of the scene. We employ 130 objects in total, gathered from 10 different kitchens, for which we collect 3D scans with a Einstar Vega 3D scanner. We built a specific apparatus for scanning objects in isolation, displayed in Figure~\ref{fig:samples}, left (for more samples, please see Appendix Fig. ~\ref{fig:mk_real_dataset_samples}). In practice, we position the object on a transparent acrylic surface. Since the scanner does not detect the acrylic surface, this allows us to take multiple scans of the same object from different viewpoints without moving it. This allows us to greatly increase the precision of object scanning. We then collect two scans per object: a first scan collected from above the object, and a second from below, in order to capture the entire 3D geometry. These two scans are later aligned to obtain a dense 3D ground truth. To facilitate alignment, we position visual markers on the acrylic surface. We use double-sided reflective markers, positioned so that they can be seen by the scanner from either above or below the object. The full process takes $6\sim10$ minutes per object.
\begin{figure}[t]
    \centering
    \includegraphics[width=\linewidth]{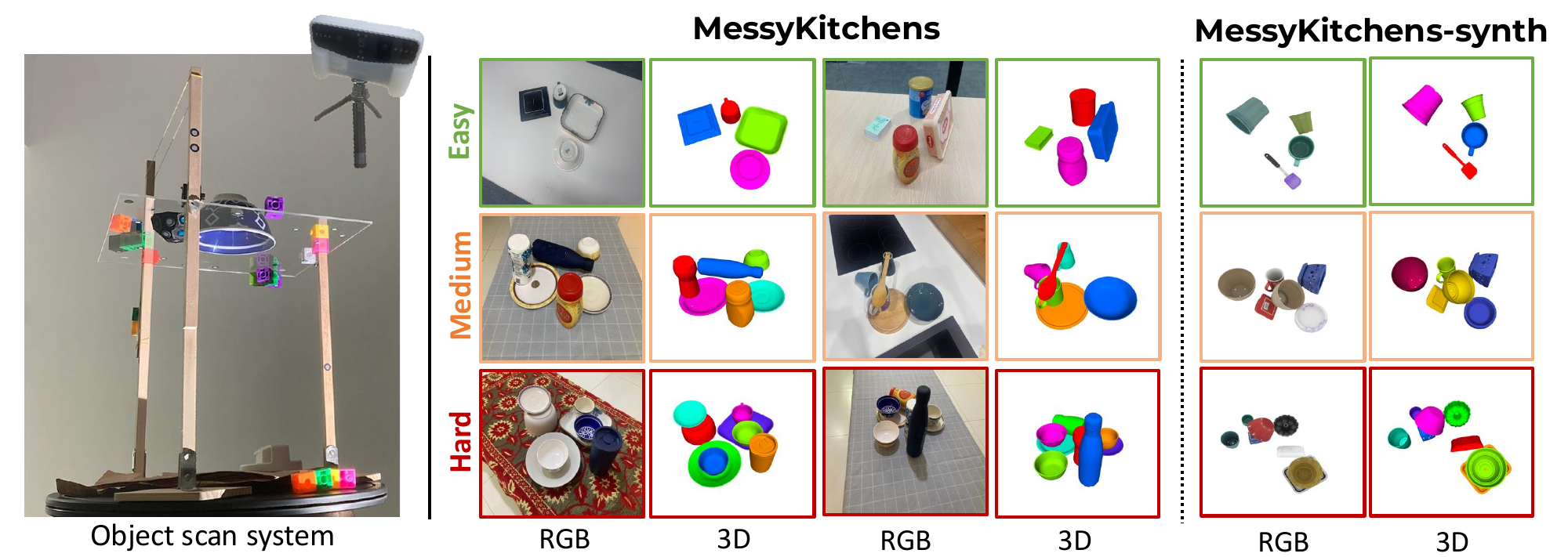}
    \caption{On the left, we show our object scanning system. The transparent surface allows us to take multiple scans without moving the object. On the right, we show samples of \benchmark, for three difficulty levels. Scenes get more cluttered and with more sophisticated object interactions with the increase of the difficulty. We also provide a synthetic set (\benchmark-synthetic) usable for training, with constructed scenes similar to the real dataset.}
    \label{fig:samples}
    \vspace{-10px}
\end{figure}

\paragraph{Difficulty levels.} We then construct a scene by assembling pre-scanned objects on different surfaces, to promote variability. For scenes, we define three different difficulty levels depending on the configuration: \textbf{(1) Easy:} we include 4 objects, well separated and with minimal contact;  \textbf{(2) Medium:} we include 6 objects, among which 4 are base objects lying on a flat surface, and the other 2 are stacked objects on top of the others, in equilibrium. We reduce the separation and impose more contacts with each other; \textbf{(3) Hard:} we use 8 objects, imposing maximum contact with each other. In addition to the characteristics of the medium setup, we include nested objects inserted one inside the other (such as a cup into a bowl). We display examples of our scenes configuration in Figure~\ref{fig:samples}, center. After being assembled, the scenes are scanned with the same sensor that was used for objects. In particular, we carefully scan each constructed setup while ensuring stable tracking and complete surface coverage, so that even heavily occluded or contact-rich regions are captured as accurately as possible. The acquired scene scans are subsequently processed, cleaned, and decimated using the scanner software, preserving geometric fidelity (with a point-to-mesh error below 0.05 mm). We subsequently map textures to the mesh. In total, the scanning process takes up to $20$ minutes per scene. The fully reconstructed scenes are exported for subsequent registration with the individual object models. Note that scenes are built incrementally: we first construct a hard scene and scan that only. We perform on the hard scene the registration of existing objects. Then, we remove some objects from the hard scene and construct, consequently, a medium scene first, and finally an easy scene. This allows us to obtain the same configuration of some objects, with varying contacts. Note that the associated RGB to the scenes is instead varying, due to the different camera pose used for recording different difficulty levels. This could be useful for assessing the dependency of 3D reconstruction algorithms from camera viewpoints (see Appendix ~\ref{sec:app_MKDataset_real} for more details on dataset creation). 

\paragraph{Registration.} For registering the object models to their corresponding scenes, we first perform a manual coarse alignment which gives us the initial transformations for our registration pipeline. Our registration pipeline is in two stages: in the first stage, we randomly sample a set of 3D points from the surface of each object mesh, find the distance to closest on-surface point on the scene mesh, and then optimize the object transformations to minimize this cost, and in the second, taking the transformations from first stage as new initial estimates, we impose, at optimization time, that the normals of the scan and of the object are aligned for a given point. Indeed, most of our objects are thin and concave, so a point on the top surface and its opposite point on the bottom surface often have very similar distances to the scan. During automatic registration, the optimizer can therefore minimize error by placing the scan surface between the two walls of the object, incorrectly satisfying both sides. Penalizing for normals coherence can prevent this (for more details, please refer Appendix ~\ref{app:registration_details}).\looseness=-1

\subsection{Synthetic data}\label{sec:method-synth}

\paragraph{Scene construction.} To enable training on \benchmark, we construct a synthetic dataset with close similarity to our setup for real scene scanning. We call this set \benchmark-synthetic. We start by selecting a total of 42 kitchenware objects from 
the 3D assets of GSO~\cite{downs2022google}. %
We then generate scenes following the same difficulty levels as in our \benchmark real scenes. For easy scenes, we simply randomly place four objects on a flat surface. For medium scenes, after randomly placing four objects, we position two additional objects on top of existing ones, and we activate gravity to make the simulation physically realistic. We make sure that objects are effectively stable on top of others before including the scene inside the dataset. For hard scenes, we include stacked objects, as in the real setup. Ensuring a realistic randomized stacking of objects in synthetic simulation is nontrivial. To do so, we first compute object volume estimates, and drop appropriately sized objects into compatible support objects. All scenes are generated using concave, mesh-based collision, which is critical for achieving realistic stacked and nested interactions. In all cases, we ensure that our scenes are contact-rich and physically realistic. We visualize samples in Figure~\ref{fig:samples}, right (for more samples, please see Appendix Fig. ~\ref{fig:mk_dataset_samples_synth}).
\begin{figure}[t]
  \centerline{
    \includegraphics[width=1.0\textwidth]{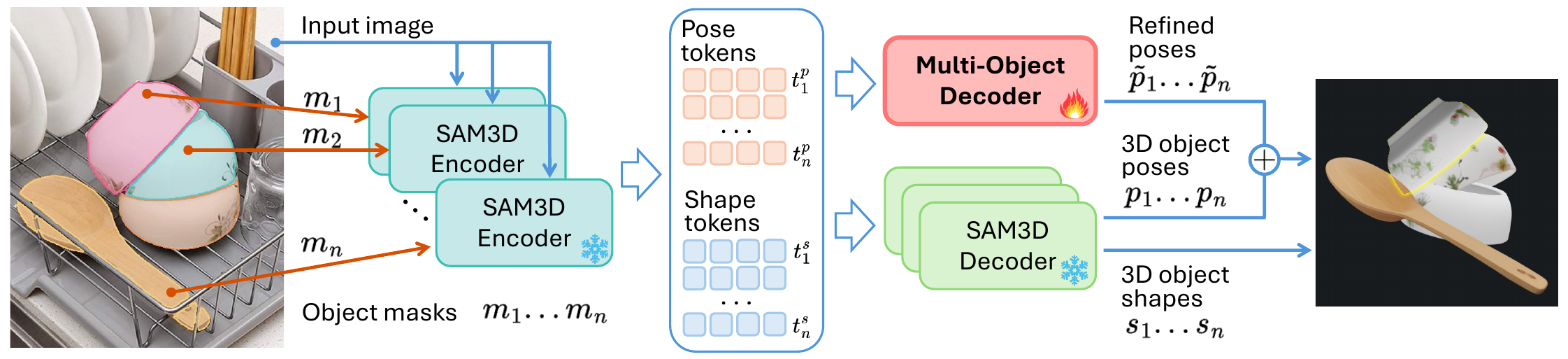}
    }
    \caption{\textbf{\methodname for 3D reconstruction.} SAM3D outputs 3D shapes from input images and masks. To impose scene-level constraints, we use a \methodname refining SAM3D prediction on the pose of the objects. The residual refined term is summed to the original prediction to obtain a scene-aware pose estimation.}
  \label{fig:method}
 \end{figure}

\paragraph{Rendering.} Once we construct the 3D scene, we sample multiple views for enabling training of 3D reconstruction methods. To promote photorealism, we employ Blender's Cycles engine for the rendering of scenes and use the original texture files from the dataset.\\

\begin{wrapfigure}{r}{0.5\textwidth} %
    \centering
    \includegraphics[width=\linewidth]{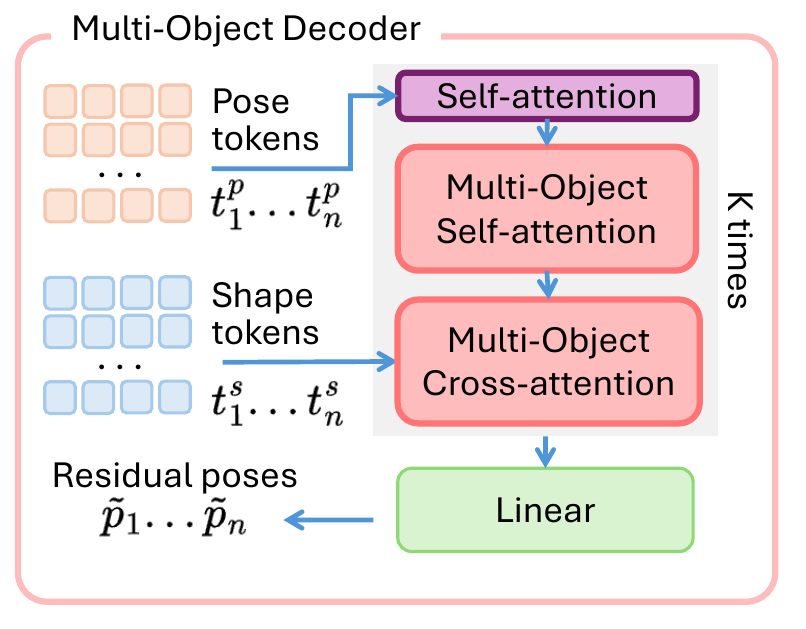}
    \caption{\textbf{\methodname.} We inform pose tokens on scene-level context by using $K$ blocks including multi-object self-attentions and cross-attention. We use pose and shape information from all objects to obtain residual pose correcting factors.} \vspace{-50px}
    \label{fig:MODecoder}
\end{wrapfigure}

We render 10 different views randomizing the camera angle, sampling from azimuth $\phi\in[0,2\pi]$ and elevation $\iota\in[\pi/4,\pi/2]$. This sampling ensures diverse viewpoints while maintaining a top-down bias consistent with tabletop scenarios. Simultaneously with the image rendering, we extract instance-based semantic maps for all objects. For more details, please see Appendix ~\ref{sec:app_MKDataset_synt}.

\section{\methodname}\label{sec:method-mod}

\subsection{Method details}
To extract 3D objects from 2D images, SAM3D takes as input an image $x$ and an object pixelwise mask $m_i$ indicating the position of object $i$. It then returns a voxel-based shape prediction $s$ and a 7-DOF pose $p = (\mathbf{q}, \mathbf{t}, \sigma)$ of the object in 3D space, where $\mathbf{q} \in \mathbb{H}_1$ is a unit quaternion representing 3D rotation, $\mathbf{t} \in \mathbb{R}^3$ is the translation vector, and $\sigma \in \mathbb{R}^+$ is the isotropic scaling factor. To do so, it processes $x$ by extracting shape tokens $t^s \in \mathbb{R}^{F_s \times C}$ and pose tokens $t^p \in \mathbb{R}^{F_p \times C}$, used to estimate respectively $s$ and $p$ through the SAM3D decoder. $F_s$ and $F_p$ are the sequence lengths, and $C$ the feature dimension. We refer to the original paper for details~\cite{chen2025sam}. For a scene with $N$ objects, we denote the aggregated scene tokens as $\mathbf{T}^s = \{t^s_1,\dots,t^s_N\} \in \mathbb{R}^{N \times F_s \times C}$ and $\mathbf{T}^p = \{t^p_1,\dots,t^p_N\} \in \mathbb{R}^{N \times F_p \times C}$, where $t^s_i$ and $t^p_i$ correspond to the shape and pose tokens of object $i$, respectively. Our objective is to train a \methodname to estimate residual scene-aware pose updates $\mathbf{\tilde{P}} = \{\tilde{p}_1,\dots,\tilde{p}_N\}$ from the aggregated tokens $\mathbf{T}^p$ and $\mathbf{T}^s$, and compute the final refined poses as $\mathbf{P} + \mathbf{\tilde{P}}$, encouraging globally consistent scene-level understanding. We visualize our approach in Figure~\ref{fig:method}. To correctly predict scene-aware pose tokens, each pose prediction should be performed with awareness of both the pose and the shape of all other objects. We achieve this objective by building \methodname as a stack of $K$ blocks composed of (i) a multi-object self-attention layer for pose tokens, which can correlate the pose of all objects, and (ii) a multi-object cross-attention that grounds the refined pose tokens to the shape tokens of all objects. As we show in Figure~\ref{fig:MODecoder}, in a block, we first refine pose tokens independently for each object by applying standard self-attention $\text{SA}(\cdot)$ along the sequence dimension, and using $N$ as a batch size:
\begin{equation}
\hat{\mathbf{T}}^p = \mathrm{SA}(\mathbf{T}^p),
\qquad 
\hat{\mathbf{T}}^p\in\mathbb{R}^{N\times F_p\times C}.
\end{equation}
This step increases the expressive capacity of pose features while preserving object-wise separation. We then enable cross-object reasoning by flattening the object and token dimensions of $\mathbf{T}^p$ into a single sequence of length $N F_p$, similarly to related literature~\cite{lin2025partcrafter,huang2025midi}, hence ${\mathrm{}}\hat{\mathbf{{T}}}^p\in\mathbb{R}^{1\times (NF_p) \times C}$ now. We then process it with a self-attention layer $\text{SA}_\text{multi}(\cdot)$:

\begin{equation}
\tilde{\mathbf{T}}^p = \text{SA}_\text{multi}(\hat{\mathbf{T}}^p), \qquad \tilde{\mathbf{T}}^p\in\mathbb{R}^{1\times (N F_p)\times C}
\end{equation}
Finally, to ground pose predictions to geometry, we similarly process shape and pose tokens with a multi-object cross-attention, using the aggregated pose tokens $\mathbf{\tilde{T}}^p$ as queries and reshaped shape tokens $\mathbf{T}^s\in \mathbb{R}^{1\times (N F_s)\times C}$ as keys and values:
\begin{equation}
\mathbf{T}_{\text{out}} 
= \mathrm{CA}_{\text{multi}}(\tilde{\mathbf{T}}^p, \mathbf{T}^s),
\qquad
\mathbf{T}_{\text{out}}\in\mathbb{R}^{1\times (N F_p)\times C}.
\end{equation}
The result $\mathbf{T}_{\text{out}}$ is finally reshaped to $\mathbb{R}^{N\times F_p\times C}$, and is provided as input for the next block. The output of the last block is decoded to $\tilde{\mathbf{P}}$ with a linear layer, independently for each refined pose.

\subsection{Training and inference}
We train \methodname with a weighted combination of rotation, translation, and scale losses. First, we impose a geometry term $\mathcal{L}_\text{CD}$ based on the Chamfer distance (CD) between the predicted and GT shapes. Then, we represent rotations as unit quaternions $\mathbf{q}\in \mathbb{H}_1$ and take advantage of an alignment term based on the predicted and ground truth quaternion rotation $\mathbf{\hat{q}}$:
$\mathcal{L}_{\mathrm{ip}} = 1-\langle \mathbf{q}, \hat{\mathbf{q}}\rangle^{2}$.
This loss accounts for the double-cover property of the quaternions, where $\mathbf{q}$ and $-\mathbf{q}$ represent the same physical rotation; by squaring the inner product, we effectively minimize the geodesic distance on the $SO(3)$ manifold without sign ambiguity~\cite{huynh2009metrics}. Translation and scale are supervised with standard regression losses between outputs (the translation vector $\mathbf{t} \in \mathbb{R}^3$ and the isotropic scaling factor $\sigma \in \mathbb{R}^+$) and the ground truth, \textit{i.e.} $\mathcal{L}_t$ and $\mathcal{L}_s$, respectively. Explicitly, the symmetric Chamfer distance is:
\begin{equation}
\mathcal{L}_{\text{CD}}(S, \hat{S}) = \frac{1}{|S|} \sum_{x \in S} \min_{\hat{x} \in \hat{S}} \|x - \hat{x}\|_2^2 + \frac{1}{|\hat{S}|} \sum_{\hat{x} \in \hat{S}} \min_{x \in S} \|x - \hat{x}\|_2^2.
\end{equation}

This geometric term is vital for handling object symmetry; since it relies on nearest-neighbor correspondence rather than fixed point-wise mapping, geometrically identical views of symmetric objects result in an equivalent loss, preventing the model from being penalized for valid but non-unique orientations. Considering that our training data and SAM 3D outputs may lie in different canonical spaces, to obtain reliable ground truth, we first estimate a global $Sim(3)$ transformation between the predicted and ground truth scenes using ICP~\cite{besl1992method}. We then match predicted objects to ground truth ones and, for each matched pair, refine the object pose by performing $Sim(3)$ ICP from the SAM 3D object to its ground truth counterpart. Finally, we decompose the resulting $Sim(3)$ transformation into rotation, translation, and scale (denoted as $\hat{\mathbf{q}}, \hat{\mathbf{t}},$ and $\hat{\sigma}$ respectively) to serve as direct supervision. Our final objective is:

\begin{equation}
\mathcal{L}
=
\lambda_{\mathrm{CD}}\mathcal{L}_{\mathrm{CD}}
+\lambda_t\mathcal{L}_t
+\lambda_s\mathcal{L}_s
+\lambda_{\mathrm{ip}}\mathcal{L}_{\mathrm{ip}}.
\end{equation}
At inference, we extract tokens for each object detected by SAM 3~\cite{carion2025sam} and refine their pose predictions with \methodname. The structural outputs of SAM 3D, such as voxels or meshes, remain unchanged but are accurately re-positioned in the 3D scene to ensure global coherence and physical consistency.

\section{Experiments}

\subsection{Setup and baselines}

\paragraph{Datasets}
We compare with multiple concurrent benchmarks including household and kitchenware objects, such as T-LESS~\cite{T-LESS}, LINEMOD~\cite{LINEMOD}, YCB-Video~\cite{YCB-V}, MP6D~\cite{MP6D}, GraspNet-1B~\cite{graspnet-1b}, GraspClutter6D~\cite{graspclutter6d}. 
For the evaluation of MOD, besides \benchmark, we use GraspNet-1B \cite{graspnet-1b}, HouseCat6D~\cite{housecat6d}, and GraspClutter6D~\cite{graspclutter6d} as an out-of-distribution test set.

\paragraph{Methods}
We compare the 3D reconstruction of \methodname against three major object-level baselines, PartCrafter~\cite{lin2025partcrafter}, MIDI~\cite{huang2025midi}, and SAM 3D~\cite{chen2025sam}. PartCrafter is the only method that does not require a segmentation map of the objects as input. For all the others and ours, we use the same segmentation map extracted by SAM 3~\cite{carion2025sam}. For all, we compare in terms of intersection-over-union (IoU) and Chamfer Distance (CD). For both metrics, we report object-level and scene-level values.

\paragraph{Training setup}
We train our \methodname on 4 NVIDIA A100 GPUs (40GB). For training, we use \benchmark-synthetic, sampling 600 scenes per difficulty level (easy/medium/hard), and rendering 6 images per scene, for a total of 10800 images. MOD adds a few parameters to SAM 3D (approximately 81 million), and training for 10 epochs takes approximately 2 hours. We set $K=3$ for the architecture of MOD. We used a learning rate of $5\times 10^{-5}$ with a linear warm-up during the first $10\%$ of training.\\

\subsection{Data quality}
\begin{figure}[t]
\centering
\begin{subtable}{0.55\linewidth}
\resizebox{\linewidth}{!}{    
\begin{tabular}{ll|ccc}
\toprule
Dataset & Sensor & $\mu_{|\delta|}$ & $\mathrm{med}_{|\delta|}$ & $\sigma_{\delta}$ \\
\midrule

\multirow{2}{*}{T-LESS \cite{T-LESS}}
  & Camarine           & 4.28  & 2.46  & 7.72  \\
  & Kinect v2          & 8.40  & 5.45  & 11.36 \\\midrule

\multirow{1}{*}{LINEMOD \cite{LINEMOD}}
  & Kinect v2          & 5.89  & 5.57  & 1.47  \\\midrule

\multirow{1}{*}{YCB-Video \cite{YCB-V}}
  & Xtion Pro Live     & 3.95  & 3.66  & 2.26  \\\midrule

\multirow{1}{*}{MP6D \cite{MP6D}}
  & Tuyang FM851-E2    & 3.54  & 2.70  & \textbf{0.17}  \\\midrule

\multirow{2}{*}{GraspNet-1B \cite{graspnet-1b}}
  & RealSense D435     & 7.69  & 4.95  & 14.30 \\
  & Azure Kinect       & 14.79 & 9.54  & 20.20 \\\midrule

\multirow{4}{*}{GraspClutter6D \cite{graspclutter6d}}
  & Zivid               & 3.22 & 1.55 & 11.10 \\
  & RealSense D415      & 5.71          & 3.77          & 14.67 \\
  & RealSense D435      & 7.02          & 4.58          & 13.82 \\
  & Azure Kinect        & 13.85         & 6.83          & 32.59 \\

\midrule
MessyKitchens
  & Einstar Vega
    & \textbf{1.62}
    & \textbf{0.91}
    & 3.83 \\

\bottomrule
\end{tabular}}
\caption{Registration accuracy}\label{tab:registration}
\end{subtable}
\hfill
\raisebox{0px}{
\begin{subfigure}{0.40\linewidth}
    \includegraphics[width=\linewidth]{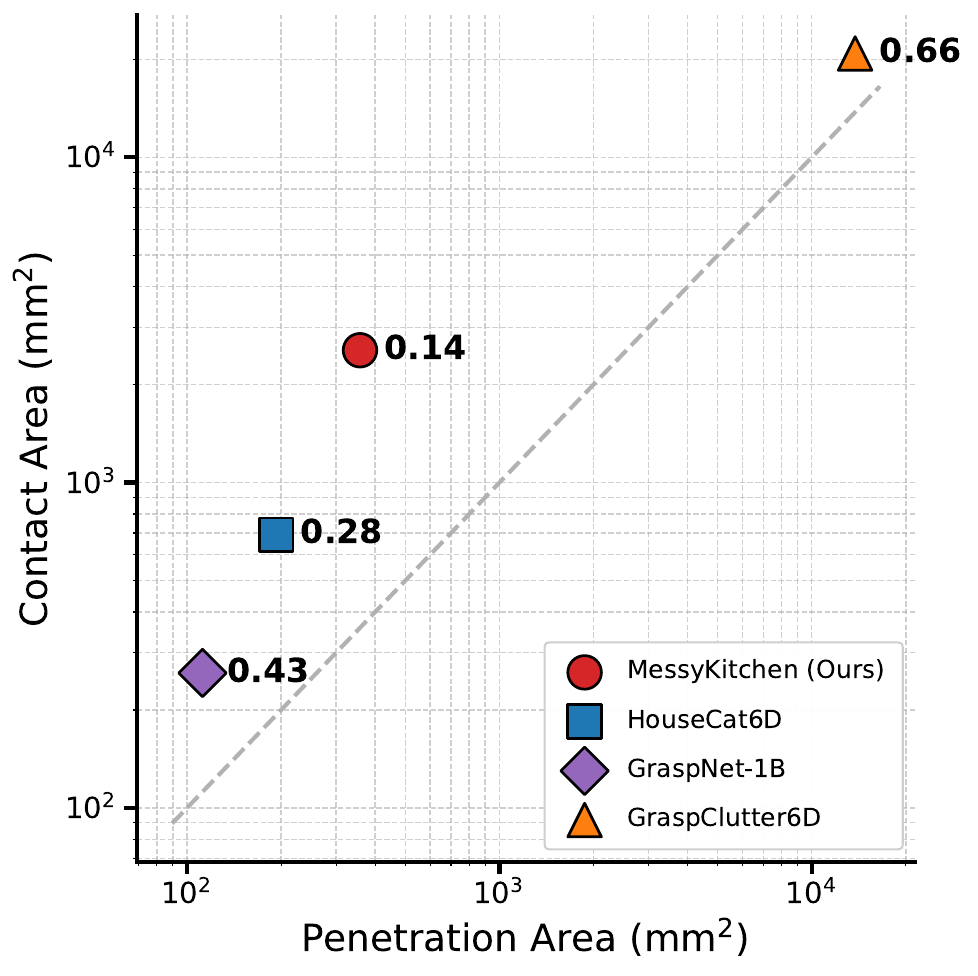}
    \caption{Contacts and penetration}\label{fig:contacts-penetration}
\end{subfigure}
}
\caption{\textbf{Comparison with other benchmarks.} In Table~\subref{tab:registration}, we show that \benchmark yields significant improvements in registration accuracy, measured with depth errors ($mm$), compared to others. In Figure~\subref{fig:contacts-penetration}, we calculate the ratio between penetration area and contacts surface area. \benchmark exhibits the best ratio, demonstrating the high quality of our cluttered scenes, resulting in physically-realistic contacts.}
\label{tab:depth_stats}

\end{figure}

\paragraph{Registration accuracy}
\begin{wraptable}{r}{0.5\linewidth}
\vspace{-35px}
\centering
\caption{\textbf{Contacts across difficulty levels.} 
We report contacts accuracy metrics across difficulty levels. The ratio between penetration and contacts in medium and hard scenes is similar, showcasing the quality of our data.}
\label{tab:reg_physics_transposed}
\setlength{\tabcolsep}{8pt}

\small
\resizebox{\linewidth}{!}{
\begin{tabular}{l|ccc}
\toprule
\multirow{2}{*}{\textbf{Split}}
& \multicolumn{3}{c}{\textbf{Contacts measures ($mm^2$)}} \\
& C. Area 
& P. Area 
& Ratio \\
\midrule

Easy   & 14.66 & 1.062 & 0.0724 \\
Medium & 2593 & 397.4 & 0.1533 \\
Hard   & 5892 & 793.5 & 0.1347 \\

\bottomrule

\end{tabular}}
\vspace{-15px}
\end{wraptable}Registration accuracy is crucial for faithful 3D reconstruction. Following~\cite{graspclutter6d,T-LESS}, we assess registration quality by measuring the depth discrepancy between the rendered depth of registered objects and the ground-truth scan depth. In Figure~\ref{tab:registration}, we report the mean ($\mu_{|\delta|}$) and median ($\mathrm{med}_{|\delta|}$) absolute depth error, together with the standard deviation $\sigma_\delta$, all in millimeters. As shown, \benchmark achieves \textit{very accurate registration}, with a mean error of \textbf{1.62} mm, corresponding to a 49.7\% relative improvement over the second-best benchmark, GraspClutter6D (3.22). We observe a similar trend for the median error, where \benchmark attains \textbf{0.91} mm, improving by 41.3\% over the second best (1.55). These results highlight the precision of our data and the effectiveness of our automatic registration pipeline, demonstrating that \benchmark provides a reliable foundation for object-level 3D reconstruction.\looseness=-1

\paragraph{Contacts and penetrations}
 Realistic modeling of contacts is essential in cluttered scenes, as many downstream tasks rely on physically plausible object interactions. However, imprecise dataset construction may introduce unrealistic inter-object penetrations, limiting the reliability of the data for contact reasoning and physics-based applications. In Figure~\ref{fig:contacts-penetration}, we compare the datasets by jointly analyzing the contact and penetration statistics. On the $y$-axis, we report the contact area, computed as the surface of points lying within 2.5mm between distinct objects. On the $x$-axis, we measure the penetration area, defined as the area of surfaces from one object intersecting another, detected via voxelization and intersection testing. Next to each point, we report the ratio between penetration surface and contact surface areas. While GraspClutter6D exhibits large contact regions, it also shows substantial penetration, resulting in an unfavorable ratio (0.66), suggesting that many contacts in the 3D scenes are not physically realistic. In contrast, \benchmark achieves the best ratio (\textbf{0.14}), showcasing the precision of our registered cluttered scenes. This serves as a further proof of the physical consistency of object interactions in \benchmark. Finally, in Table~\ref{tab:reg_physics_transposed}, we report ratios between contacts and penetration in easy/medium/hard scenes. We notice that the ratio obtained in medium/hard scenes is similar, highlighting that our registration is robust enough for physically-accurate contacts independently from the complexity. Easy scenes have no contacts.

\begin{table*}[t]
\centering
\caption{\textbf{Effectiveness of \methodname.} On \benchmark, GraspNet-1B, HouseCat6D, and GraspClutter6D, SAM 3D+MOD achieves state-of-the-art object-level 3D scene reconstruction. We measure both object-level metrics and scene-level ones. GraspClutter6D~\cite{graspclutter6d} is a challenging out-of-distribution benchmark with high clutter; all MOD evaluations are zero-shot.}
\label{tab:recon_comparison_all}
\setlength{\tabcolsep}{6pt}
\small
\resizebox{1.0\textwidth}{!}{%
\begin{tabular}{c lcc|cc|cc|cc}
\toprule
& & \multicolumn{2}{c|}{\benchmark} & \multicolumn{2}{c|}{GraspNet-1B} & \multicolumn{2}{c|}{HouseCat6D} & \multicolumn{2}{c}{GraspClutter6D} \\
\cmidrule(lr){3-4}\cmidrule(lr){5-6}\cmidrule(lr){7-8}\cmidrule(lr){9-10}
& \textbf{Method}
& IoU$\uparrow$ & CD$\downarrow$
& IoU$\uparrow$ & CD$\downarrow$
& IoU$\uparrow$ & CD$\downarrow$
& IoU$\uparrow$ & CD$\downarrow$ \\
\midrule
\multirow{4}{*}{\rotatebox{90}{\textbf{Object}}}
& PartCrafter & 0.071 & 0.495 & 0.020 & 0.956 & 0.029 & 0.852 & 0.067 & 0.674 \\
& MIDI     & 0.186 & 0.285 & 0.067 & 0.640 & 0.092 & 0.433 & 0.086 & 0.500 \\
& SAM 3D        & 0.409 & 0.064 & 0.336 & 0.082 & 0.325 & 0.125 & 0.328 & 0.105 \\
& \textbf{MOD} & \textbf{0.445} & \textbf{0.061} & \textbf{0.344} & \textbf{0.078} & \textbf{0.404} & \textbf{0.100} & \textbf{0.340} & \textbf{0.103} \\
\midrule
\multirow{4}{*}{\rotatebox{90}{\textbf{Scene}}}
& PartCrafter & 0.133 & 0.228 & 0.075 & 0.355 & 0.114 & 0.289 & 0.227 & 0.227 \\
& MIDI     & 0.238 & 0.165 & 0.121 & 0.327 & 0.172 & 0.215 & 0.213 & 0.201 \\
& SAM 3D        & 0.431 & 0.054 & 0.356 & 0.074 & 0.374 & 0.099 & 0.487 & 0.059 \\
& \textbf{MOD} & \textbf{0.472} & \textbf{0.050} & \textbf{0.377} & \textbf{0.069} & \textbf{0.458} & \textbf{0.079} & \textbf{0.496} & \textbf{0.058} \\
\bottomrule
\end{tabular}
}
\vspace{-15px} %
\end{table*}

\begin{figure}[t]
    \centering
    \includegraphics[width=\linewidth]{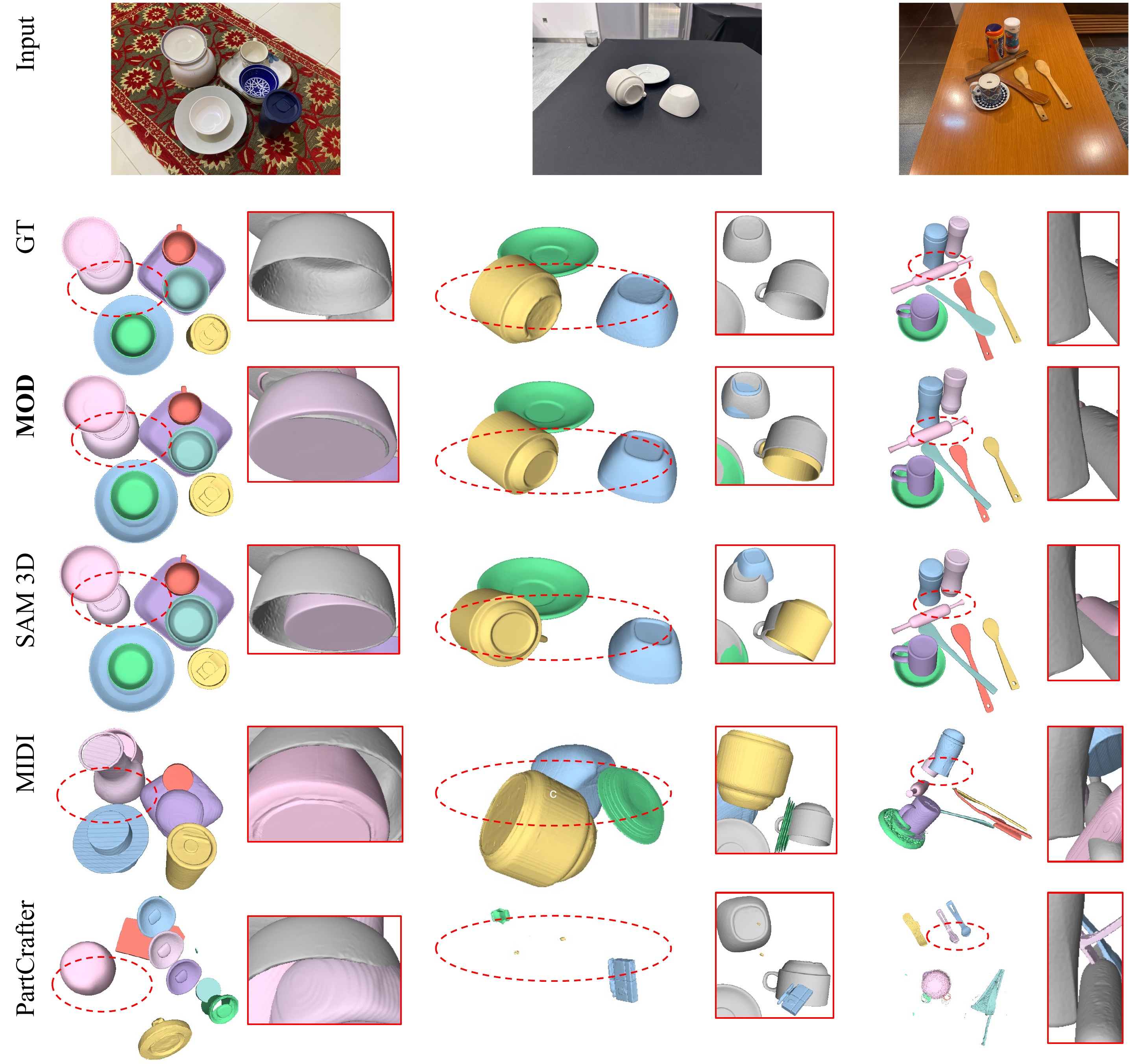}
    \caption{\textbf{Qualitative comparison.} We show examples of 3D reconstructions for MOD and alternative methods, on \benchmark. In the insets, we show significant scene-level improvements over SAM 3D, demonstrating the effectiveness of MOD. The gray shapes are ground truth.\looseness=-1 }
    
    \label{fig:qualitative_comparison}
\vspace{-20px}
\end{figure}

\begin{figure}[t]
    \centering
    \includegraphics[width=\linewidth]{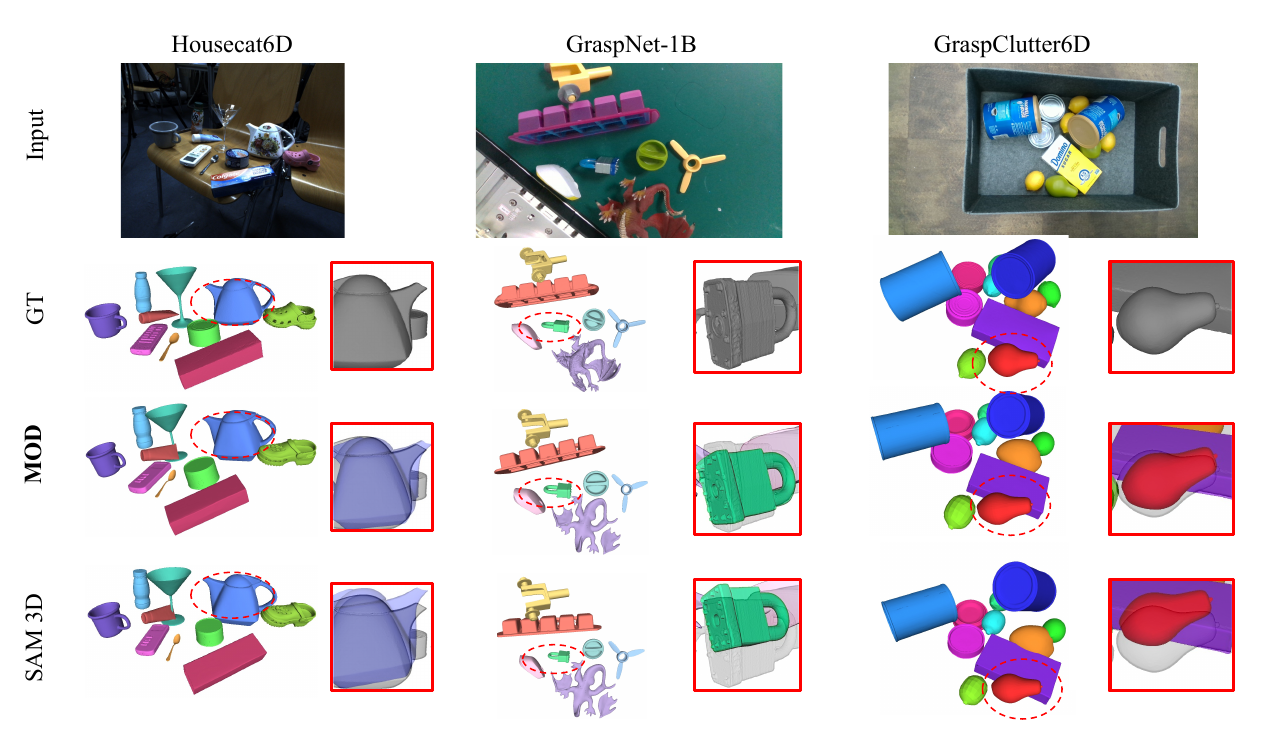}
    \caption{\textbf{Qualitative comparison across Housecat6D, GraspNet-1B, and GraspClutter6D.} We visualize 3D scene reconstructions for MOD and the SAM 3D baseline. Our method consistently produces physically more plausible results with fewer interpenetration and more accurate grounding compared to the baseline. Ground-truth meshes are overlaid in gray with transparency for reference.\looseness=-1}
\label{fig:qualitative_ood}
\vspace{-15px}
\end{figure}

\subsection{3D object-based reconstruction}
\paragraph{Effectiveness of MOD}

In Table~\ref{tab:recon_comparison_all}, we compare \methodname against state-of-the-art baselines for object-centric 3D reconstruction. We train \methodname exclusively on \benchmark-Synthetic, while evaluating SAM 3D, PartCrafter, and MIDI in a zero-shot setting. 
Evaluation is conducted on \benchmark together with three out-of-distribution benchmarks, GraspNet-1B~\cite{graspnet-1b}, HouseCat6D~\cite{housecat6d}, and GraspClutter6D~\cite{graspclutter6d}, on which all \methodname{} evaluations are zero-shot. %
As predicted 3D reconstructions and ground-truth scenes may not share a common coordinate frame, we perform scene-level alignment using Sim(3) ICP. To mitigate sensitivity to initialization and avoid alignment bias, we run ICP three times and report the best result. Naive inference with SAM 3D already substantially outperforms competing methods across all benchmarks, achieving, for example, a +19.3\% improvement over the second best (MIDI) in scene-level reconstruction (0.238 vs 0.431), motivating our design choice to use \methodname as a plug-in component to SAM 3D. However, incorporating \methodname further boosts performance consistently. For instance, for object-level IoU, training solely on synthetic data yields improvements on \benchmark (0.409 vs \textbf{0.445} with \methodname), as well as gains on GraspNet-1B (0.336 vs \textbf{0.344}), HouseCat6D (0.325 vs \textbf{0.404}), and GraspClutter6D (0.328 vs \textbf{0.340}). These results indicate that \benchmark provides effective supervision for robust object-based 3D reconstruction and highlight the importance of accurate contacts and object pose/scale refinement, as even minor geometric corrections lead to notable improvements. Furthermore, \methodname demonstrates strong generalization under domain shift: HouseCat6D, GraspNet-1B, and GraspClutter6D contain object categories distinct from kitchenware, confirming OOD generalization to different objects. %

\begin{wraptable}{r}{0.35\linewidth}
\vspace{-30px}
\centering
\caption{\textbf{Contact accuracy.} We build contact graphs as~\cite{ym2025physic} to evaluate the quality of MOD-reconstructed contacts. }
\label{tab:contact_metrics}
\setlength{\tabcolsep}{4pt}
\resizebox{\linewidth}{!}{
\begin{tabular}{l|ccc}
\toprule
\multirow{2}{*}{\textbf{Method}}
& \multicolumn{3}{c}{\textbf{Contact (\%)}} \\
& Prec$\uparrow$
& Rec$\uparrow$
& F1$\uparrow$ \\
\midrule
MIDI        & 38.9 & 48.1 & 43.0 \\
PartCrafter & 27.2 & 17.4 & 21.2 \\
SAM 3D      & 66.0 & 48.3 & 55.8 \\
\textbf{MOD} & \textbf{70.0} & \textbf{62.5} & \textbf{66.1} \\
\bottomrule
\end{tabular}}
\vspace{-38px}
\end{wraptable}

Finally, in Table~\ref{tab:contact_metrics} we quantify the physical plausibility of the reconstructions on MK using the contact metrics of PhySIC~\cite{ym2025physic}, built from a graph linking objects in contact. MOD ranks first, showing that our contribution improves contact quality. We report per-difficulty analysis of MOD on MessyKitchens dataset in Appendix ~\ref{sec:app_per_difficulty_analysis}.

\paragraph{Qualitative results}
We report in Figure~\ref{fig:qualitative_comparison} a qualitative comparison between \methodname and alternative methods on \benchmark. As shown, PartCrafter and MIDI struggle to generalize to the evaluated setups, often producing incomplete or geometrically inconsistent reconstructions. SAM 3D instead generates visually realistic object shapes across \benchmark and the other datasets. \methodname builds on these predictions by refining only the pose and scale of the detected objects, leaving their geometry unchanged. In the insets, we present close-up comparisons between SAM 3D and \methodname outputs (ground truth shown as transparent gray surfaces). The addition of \methodname consistently improves alignment in the presence of object interactions, correcting pose and scale estimates by enforcing scene-level geometric consistency and more accurate inter-object spatial relationships. In particular, the last column shows how \methodname correctly reconstructs two objects in contact with each other, thanks to our precise contact-rich training data in \benchmark.
We further report qualitative reconstructions on the three out-of-distribution (OOD) benchmarks in Figure~\ref{fig:qualitative_ood}: GraspNet-1B, HouseCat6D, and GraspClutter6D. As shown, \methodname demonstrates superior generalization compared to SAM 3D. Despite the significant domain gap between our synthetic training data and these real-world captures, \methodname maintains global scene coherence, resolving common artifacts such as inter-object penetrations and unstable ``floating'' poses that frequently occur in the baseline reconstructions. This highlights the effectiveness of our learned object-level priors in reasoning about complex, contact-rich interactions across diverse environments.

\subsection{Ablation studies}
\begin{figure}[t]
\begin{subfigure}{0.3\linewidth}
\includegraphics[width=\linewidth]{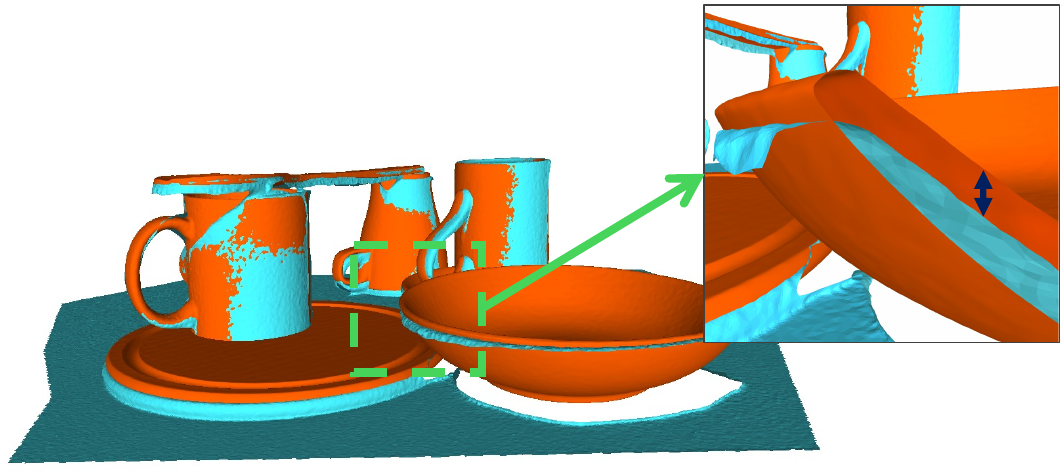}    
\caption{Distance only}\label{fig:d_only}
\end{subfigure}
\begin{subfigure}{0.3\linewidth}
    \includegraphics[width=\linewidth]{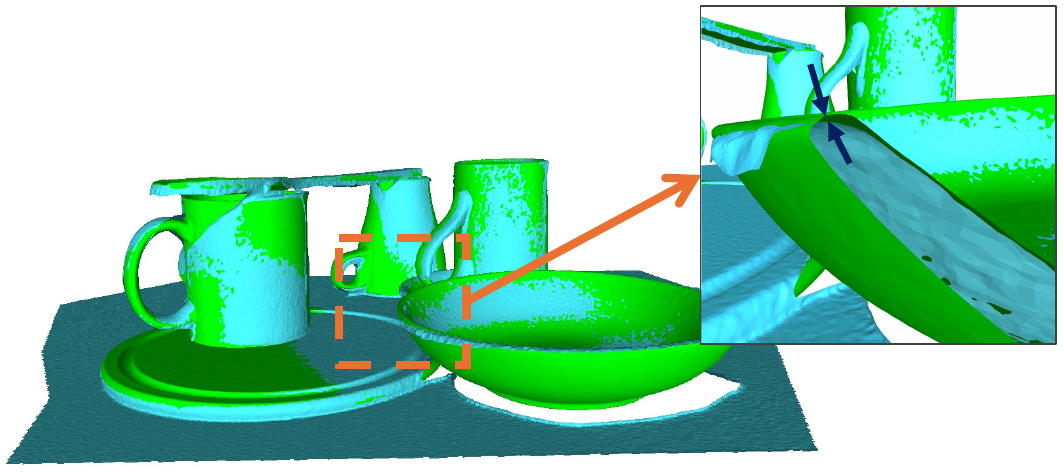}
    \caption{Distance+normals}\label{fig:d+n}
\end{subfigure}
\begin{subtable}{0.35\linewidth}
   \resizebox{\linewidth}{!}{
    \begin{tabular}{cccc}
    \toprule
         \textbf{Registration} & $\mu_{|\delta|}$ & $\text{med}_{|\delta|}$ & $\sigma_\delta$\\
         \midrule
         Manual & 4.689 &  3.209 & 7.439\\
         Distance only & 2.892 & 2.141 & 4.822 \\
         Distance+normals & \textbf{1.615} & \textbf{0.911} & \textbf{3.827} \\
         \bottomrule
         & 
    \end{tabular}}\vspace{-10px}
    \caption{Quantitative}\label{tab:registration_quant-abl}
\end{subtable}
    \caption{\textbf{Registration ablation.} We show that our approach based on distance+normal (Fig.~\subref{fig:d+n}) registration considerably improves results with respect distance only (Fig.~\subref{fig:d_only}). Metrics on registration (Table~\subref{tab:registration_quant-abl}) agree with our evaluation.}\label{fig:ablation-registration} 
    \vspace{-15px}
\end{figure}

\paragraph{Registration strategy}
We ablate our design for the registration strategy, following the same metrics reported in Figure~\ref{tab:registration}. First, we show %
the rough manual initialization used for our registration. Then, we perform our automatic registration, with a naive distance criteria or our matching criteria based on normals described in Sec.~\ref{sec:method-real}. From results in Figure~\ref{fig:ablation-registration}, we outperform considerably the alternatives. This shows the effectiveness of our automatic registration pipeline.

\paragraph{Multi-object attention}
In \methodname, we use multi-object self- and cross- attention, for pose and shape tokens. Now, we investigate the effectiveness of our choice. We compare our design (S+P) using both multi-object attention in shape (S) and pose (P) tokens with alternative designs in which we use multi-object attention for shape or pose only. In these, we replace the multi-object attention with a standard cross- or self-attention, operating on a single object. In Table~\ref{tab:ablation_architecture_attn}, we show that our setup achieves the best results, suggesting that the interaction between pose and shape tokens for all objects is important.\looseness=-1

\paragraph{Architecture}
We propose an ablation on the blocks used for the \methodname. In Table~\ref{tab:ablation_num_layers}, we variate $K$, the number of blocks, in the range $K=\{1,3,6\}$. From our results, $K=3$ yields the best results, while the addition of more blocks (K=6) tends to degrade results. We hypothesize that this is due to the richness of the features of SAM 3D, where we plug MOD on. Indeed, it may be that SAM 3D does not necessitate many non-linearities and complex architectures to encode scene awareness, benefiting from a smaller MOD. 

\begin{table*}[t]
\centering
\caption{\textbf{MOD ablation studies.} In Table~\subref{tab:ablation_architecture_attn}, we investigate the effects of the multi-object attention layers, showing that both scene-level information about pose and shape contribute to best performance. In Table~\subref{tab:ablation_num_layers}, we report that 3 transformer blocks to build MOD ($K=3$) yield the best results.}
\label{tab:ablation_combined}
\setlength{\tabcolsep}{5pt}
\small

\begin{subtable}[t]{0.45\textwidth}
\centering

\begin{tabular}{cl cc}
\toprule
& & \multicolumn{2}{c}{MessyKitchens} \\
\cmidrule(lr){3-4}
& \textbf{MOD Setup} & IoU$\uparrow$ & CD$\downarrow$ \\
\midrule
\multirow{3}{*}{\rotatebox{90}{\textbf{Object}}} 
& Shape only & 0.438 & 0.063 \\
& Pose only & 0.432 & 0.065 \\
& \textbf{S+P (Ours)} & \textbf{0.445} & \textbf{0.061} \\
\midrule
\multirow{3}{*}{\rotatebox{90}{\textbf{Scene}}} 
& Shape only & 0.463 & 0.054 \\
& Pose only & 0.457 & 0.056 \\
& \textbf{S+P (Ours)} & \textbf{0.472} & \textbf{0.050} \\
\bottomrule
\end{tabular}
\caption{Multi-object attention}
\label{tab:ablation_architecture_attn}
\end{subtable}
\hfill
\begin{subtable}[t]{0.45\textwidth}
\centering
\begin{tabular}{cl cc}
\toprule
& & \multicolumn{2}{c}{MessyKitchens} \\
\cmidrule(lr){3-4}
& \textbf{Blocks Setup} & IoU$\uparrow$ & CD$\downarrow$ \\
\midrule
\multirow{3}{*}{\rotatebox{90}{\textbf{Object}}} 
& $K=1$ & 0.441 & 0.061 \\
& \textbf{$K=3$ (Ours)} & \textbf{0.445} & \textbf{0.061} \\
& $K=6$ & 0.403 & 0.065 \\
\midrule
\multirow{3}{*}{\rotatebox{90}{\textbf{Scene}}} 
& $K=1$ & 0.467 & 0.051 \\
& \textbf{$K=3$ (Ours)} & \textbf{0.472} & \textbf{0.050} \\
& $K=6$ & 0.427 & 0.059 \\
\bottomrule
\end{tabular}
\caption{Number of blocks}
\label{tab:ablation_num_layers}
\end{subtable}
\vspace{-25px} %
\end{table*}

\section{Conclusion}

In this work, we addressed the challenging problem of physically-plausible, object-level 3D scene reconstruction from single images. To overcome the limitations of existing datasets, we introduced \benchmark, a novel benchmark featuring cluttered, contact-rich real-world scenes. By utilizing a rigorous data acquisition and normals-aware registration pipeline, \benchmark provides high-fidelity 3D ground truth with significantly lower inter-object penetration compared to previous datasets, setting a new standard for evaluating physical consistency. Furthermore, we proposed the MOD (\methodname), a simple yet highly effective extension for the single-object SAM 3D framework that combines several multi-object attentions.
Extensive experiments demonstrate that our approach significantly outperforms state-of-the-art baselines on \benchmark, GraspNet-1B, HouseCat6D, and GraspClutter6D showing strong out-of-distribution generalization and high-quality results. Ultimately, we believe that \benchmark and MOD will provide a robust foundation for future research in physics-consistent 3D computer vision, facilitating advancements in downstream applications such as robotic manipulation, virtual reality, and 3D animation.

\section*{Acknowledgment}
This work was supported by the Institute of Information \& Communications Technology Planning \& Evaluation (IITP) grant funded by the Korean Government (MSIT) (No. RS-2024-00457882, National AI Research Lab Project). 

We would also like to thank Komal Kumar, Mohamad Kassab, Gustavo Stahl, Ramazan Fazylov, David Romero, MBZUAI Cafe and Canteen, Amog Rao, Dinura Dissanayake and Mingfei Han for helping us with their objects.

\bibliographystyle{splncs04}
\bibliography{main}
\newpage
\section*{Appendix}

In this appendix, we first report MOD's per-difficulty performance on the easy, medium, and hard splits of MessyKitchens in Section~\ref{sec:app_per_difficulty_analysis}. We then describe the details for the creation of the MessyKitchens dataset and the synthetic trainig dataset -- MessyKitchens-synthetic in Sections~\ref{sec:app_MKDataset_real} and ~\ref{sec:app_MKDataset_synt}, respectively. The MessyKitchens dataset summary and comparison with other datasets is discussed in Section ~\ref{sec:app_MKDataset_summary}.

\appendix

\section{Per-difficulty performance on MessyKitchens}
\label{sec:app_per_difficulty_analysis}
We additionally report MOD's performance across the easy, medium, and hard splits of MessyKitchens (Table~\ref{tab:per_difficulty}). Although the hard split is more challenging \textit{by construction}---its scenes contain more objects and more contacts---we observe that MOD's accuracy does not degrade accordingly, and in fact improves on most metrics. This is expected: MOD leverages inter-object relationships to refine each object's pose, so a larger number of objects yields a richer scene representation that offsets the added difficulty.

\begin{table}[ht]
\centering
\caption{Per-difficulty performance on MessyKitchens. We report scene-level (s-) and object-level (o-) Chamfer Distance (CD) and IoU across the easy, medium, and hard splits.}
\label{tab:per_difficulty}
\setlength{\tabcolsep}{10pt}
\small
\resizebox{0.65\textwidth}{!}{
\begin{tabular}{l cccc}
\toprule
Split & s-CD$\downarrow$ & s-IoU$\uparrow$ & o-CD$\downarrow$ & o-IoU$\uparrow$ \\
\midrule
Easy   & 0.0586 & 0.4591 & 0.0768 & \textbf{0.4523} \\
Medium & 0.0457 & 0.4720 & 0.0698 & 0.4376 \\
Hard   & \textbf{0.0439} & \textbf{0.4850} & \textbf{0.0507} & 0.4421 \\
\bottomrule
\end{tabular}
}
\end{table}

\section{MessyKitchens dataset creation}
\label{sec:app_MKDataset_real}
\begin{figure*}[t]
    \centering
    \includegraphics[width=\textwidth]{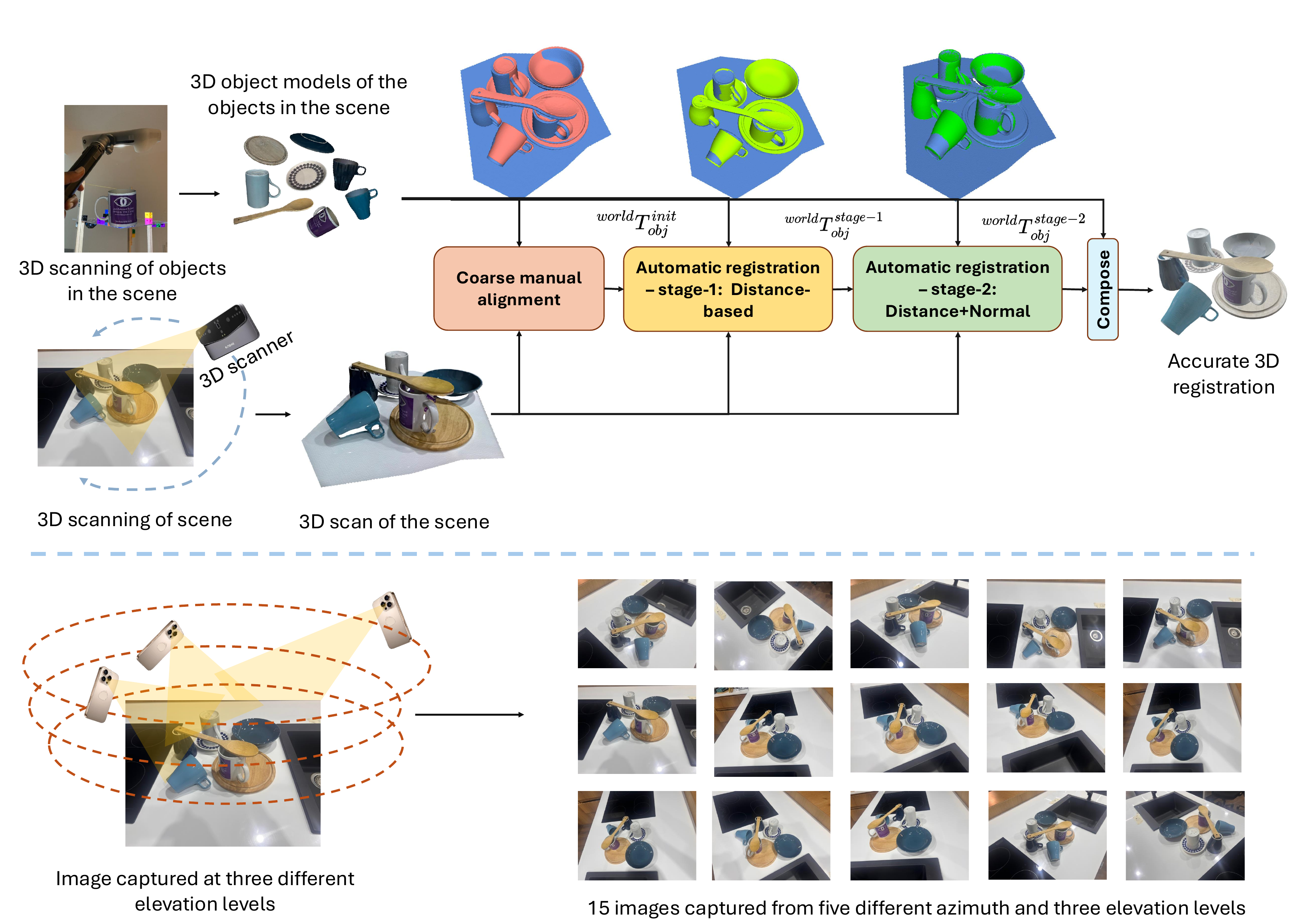}
    \caption{
    Overview of our MessyKitchens dataset acquisition and registration pipeline.
    }
    \label{fig:mk_real_dataset_creation}
\end{figure*}

In this section, we elaborate on our MessyKitchens dataset creation pipeline, see Figure~\ref{fig:mk_real_dataset_creation}.

\subsection{Object Scanning System}

Our object scanning system consists of an acrylic plate mounted on a manual turntable. 
Since a complete 3D model requires capturing both visible and occluded surfaces, each object must be scanned from at least two sides (typically top and bottom) and the resulting scans must be aligned. 
Standard scan alignment requires selecting at least three corresponding features across scans, which is tedious and often infeasible for texture-less objects. 
To address this limitation, we designed a simple yet effective solution: we attach reflective markers to an acrylic base using double-sided placement (two single-sided markers carefully aligned back-to-back so that their centers coincide). 
Because the 3D scanner can see through acrylic, the same marker is visible in both top and bottom scans, enabling straightforward manual correspondence selection in the Einstar StarVision software. 
Additionally, we mount Lego blocks with embedded 3D markers onto the base, introducing non-coplanar constraints that improve alignment stability and prevent degeneracies caused by relying solely on in-plane features. 
The turntable further accelerates scanning by exposing all object sides without moving the scanner.

\subsection{Object Scanning Procedure}

Before scanning scenes, we first create high-quality 3D models of all objects that will appear in the scenes. 
Each object is placed on the acrylic platform mounted on the turntable, and two scans are captured: one from a top-down viewpoint and one from a bottom-up viewpoint. 
To expedite acquisition, the scanner remains fixed at one of the predefined positions while the turntable is rotated, ensuring full surface coverage. 
Scanning is performed in the device's \emph{marker-only alignment} mode, where reflective markers are used to align frames within each scanning session. 
The two complementary scans are then merged in the Einstar StarVision software by manually selecting corresponding markers across the scans. 
Because the reflective markers are double-sided and visible through the acrylic base, identifying consistent correspondences between the two scans is straightforward and reliable.

\subsection{Scene Scanning}

After all objects are successfully scanned, we construct scenes using these objects. 
Although object scanning has been consistently successful, in a few cases we repeated the scanning and alignment process multiple times to obtain the highest-quality model. 
Scenes are scanned using the device's \emph{Fast} mode at the highest available resolution, with both \emph{feature} and \emph{texture} alignment enabled. 
The resulting point cloud is imported into the Einstar StarVision software, where it is reconstructed into a mesh. 
The raw scene meshes typically contain on the order of $3 \times 10^6$ vertices. 
To keep the mesh size manageable for our registration pipeline, we decimate the mesh by $80\%$. 
We verified geometric fidelity by measuring point-to-mesh error between the original and decimated meshes, observing an error below $0.05$\,mm, confirming negligible loss of accuracy.

\subsection{RGB Image Acquisition}

After completing the 3D scene scanning, we capture RGB images of each scene using a handheld cell phone camera operating in auto-focus mode. 
For each scene, we acquire 15 images in total: images are captured from three different elevation levels, and at each elevation level, five images are taken uniformly across azimuth angles to ensure diverse viewpoints. 
To increase variability and improve robustness for downstream tasks, we intentionally capture scenes under varying background settings and lighting conditions. 
This includes changes in ambient illumination and background appearance, resulting in realistic variability across the RGB image set.

\subsection{Object-to-Scene Registration}
\label{app:registration_details}
\paragraph{Distance-Based Registration.}
After manual coarse alignment, we refine each object pose by minimizing a robust surface-to-surface distance objective. Let 
$^{scene}{}T_{obj} \in SE(3)$ denote the rigid transformation of the object. 
We uniformly sample $M = 500$ points $\{\mathbf{p}_i\}_{i=1}^M$ from the objects' surface. 
For each transformed point $^{scene}{}T_{obj}\mathbf{p}_i$, we compute its closest point on the scene mesh, denoted by 
$\Pi_{\mathcal{S}}(^{scene}{}T_{obj}\mathbf{p}_i)$. 
Closest-point queries are performed using the \texttt{proximity} module of Trimesh  \cite{trimesh} library, which efficiently handles large scene meshes ($\sim$600K faces) via spatial acceleration structures.

The geometric residual for each sampled point is defined as
\begin{equation}
r_i(^{scene}{}T_{obj}) = \left\| ^{scene}{}T_{obj}\mathbf{p}_i - \Pi_{\mathcal{S}}(^{scene}{}T_{obj}\mathbf{p}_i) \right\|_2.
\end{equation}

We optimize the object transformation by minimizing the robustified least-squares objective
\begin{equation}
E_{\text{dist}}(^{scene}{}T_{obj}) = \sum_{i=1}^{N} \rho\big(r_i(^{scene}{}T_{obj})^2\big),
\end{equation}
where $\rho(\cdot)$ is the soft-$\ell_1$ loss used in SciPy's \texttt{least\_squares} solver \cite{virtanen2020scipy}:
\begin{equation}
\rho(s) = 2 \left( \sqrt{1 + \frac{s}{f^2}} - 1 \right),
\end{equation}
with $s = r_i(^{scene}{}T_{obj})^2$ and $f = 4.5$ the f-scale parameter. 
This loss behaves quadratically for small residuals and approximately linearly for large residuals, providing robustness to outliers caused by occlusions, scan noise, or missing geometry. 
We run the solver for 20 iterations in this stage.

\paragraph{Normal-Aware Registration.}
For thin and concave objects, purely distance-based alignment may produce ambiguous solutions. Points sampled from opposite sides of a thin surface may have similar distances to the scan, allowing the optimizer to incorrectly place the scene surface between two object walls while still minimizing the distance cost.
To mitigate this issue, we incorporate surface normal consistency in a second optimization stage.

For each sampled object point $\mathbf{p}_i$, we compute its surface normal $\mathbf{n}_i^{\text{obj}}$ and the normal of the scene surface at the closest point, denoted $\mathbf{n}_i^{\text{scene}}$. 
We define a normal-consistency weight
\begin{equation}
w_i =
\begin{cases}
\mathbf{n}_i^{\text{obj}} \cdot \mathbf{n}_i^{\text{scene}}, 
& \text{if } \mathbf{n}_i^{\text{obj}} \cdot \mathbf{n}_i^{\text{scene}} \ge 0.7, \\
0, & \text{otherwise}.
\end{cases}
\end{equation}

The weighted residual becomes
\begin{equation}
\tilde{r}_i(^{scene}{}T_{obj}) = w_i \, r_i(^{scene}{}T_{obj}),
\end{equation}
and the optimized objective is
\begin{equation}
E_{\text{normal}}(^{scene}{}T_{obj}) = \sum_{i=1}^{M} \rho\big(\tilde{r}_i(^{scene}{}T_{obj})^2\big).
\end{equation}

This stage is initialized with the transformation obtained from the discussed distance-only stage and optimized for 20 iterations using the same sampling strategy ($M=500$) and soft-$\ell_1$ loss.

The two-stage procedure first ensures geometric proximity and then resolves thin-surface ambiguities through normal coherence, resulting in physically plausible and stable object-to-scene alignments.

\begin{figure*}[t]
    \centering
    \includegraphics[width=\textwidth]{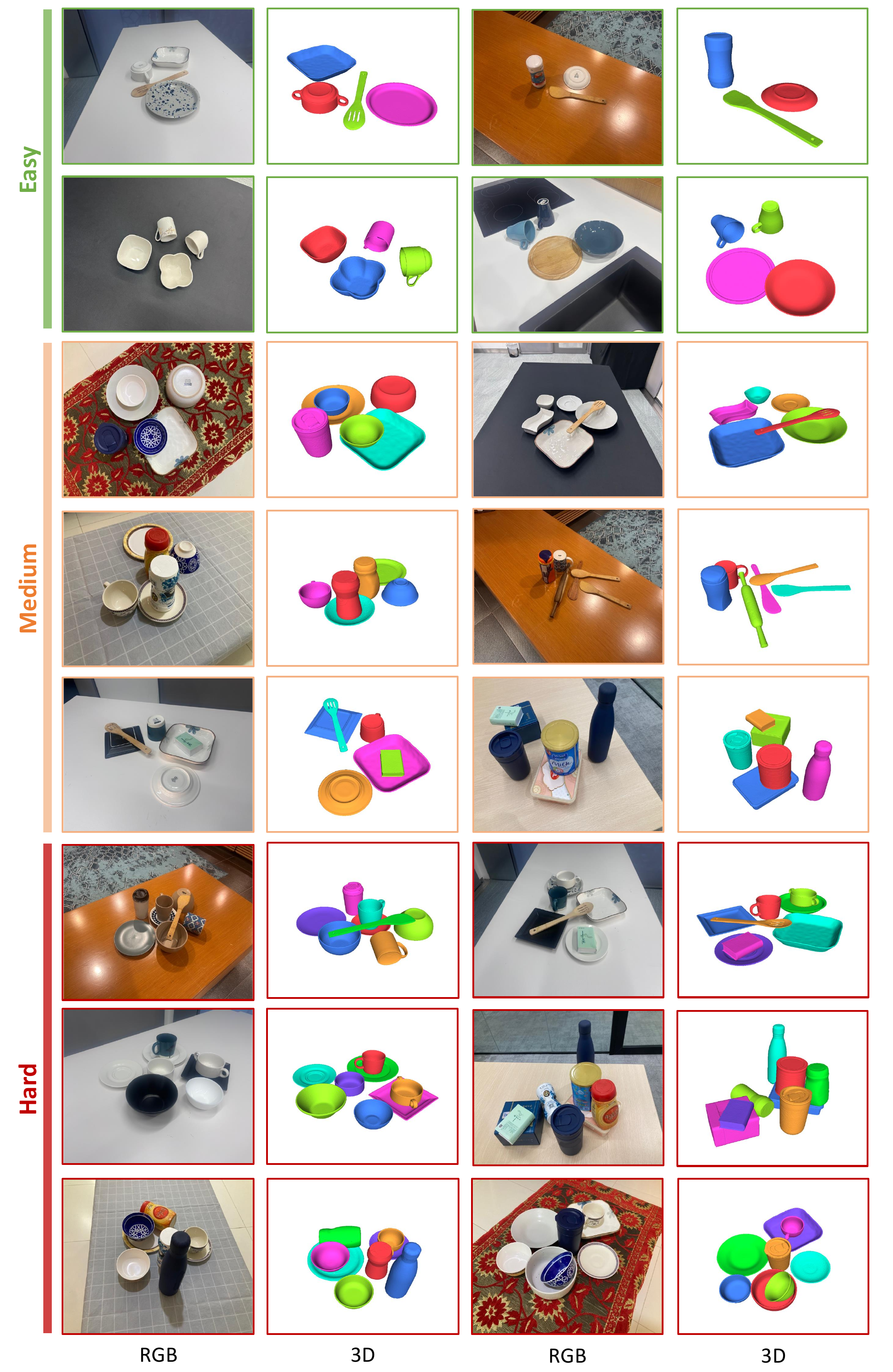}
    \caption{
    A few samples from all the three difficulty levels of our MessyKitchens %
    dataset. We show the RGB image and the ground-truth 3D acquired from our registration pipeline.
    }
    \label{fig:mk_real_dataset_samples}
\end{figure*}

In Figure \ref{fig:mk_real_dataset_samples} we show a few samples of our MessyKitchens dataset from all the three difficulty level categories -- Easy, Medium and Hard (more visualizations are  available at our project website  \url{https://messykitchens.github.io/}).  %

\section{MessyKitchens-Synthetic dataset creation}
\label{sec:app_MKDataset_synt}

\subsection{Preprocessing}

To enable training on MessyKitchens dataset, we construct a synthetic dataset closely matching the real-world acquisition setup described in the main paper. We refer to this dataset as MessyKitchens-Synthetic. We begin with the 3D object assets from GSO [18], selecting 42 kitchenware objects consistent with our real scenes. Prior to scene generation, we preprocess all meshes to ensure simulation stability. Specifically, we (1) remesh the objects to obtain well-conditioned, uniformly distributed triangle meshes suitable for physics simulation, and (2) adjust the center of mass when necessary to ensure physically correct behavior during stacking and dropping. These preprocessing steps are critical for avoiding unstable or biased object poses during simulation.

\subsection{Simulation parameters}

All scenes are simulated in Blender  using concave, mesh-based collision, which are essential for accurately modeling nested and stacked interactions. To ensure physical plausibility, we carefully tune simulation parameters. The object-object collision margin is set to 0.01mm, while the object-plane collision margin is set to 1.0mm. We disable velocity-based deactivation, as it can prematurely freeze objects in physically implausible poses. The restitution parameter is set to zero to eliminate bounce during collisions, preventing unrealistic motion. Additionally, we found that modeling the flat plane as an ``active’’ object in the physics engine, while fully constraining its translation and rotation (effectively making it passive), significantly reduces interpenetration artifacts compared to treating it as a purely static collider. All generated scenes are contact-rich, stable, and visually consistent with real-world stacking behavior.

\begin{figure*}[t]
    \centering
    \includegraphics[width=\textwidth]{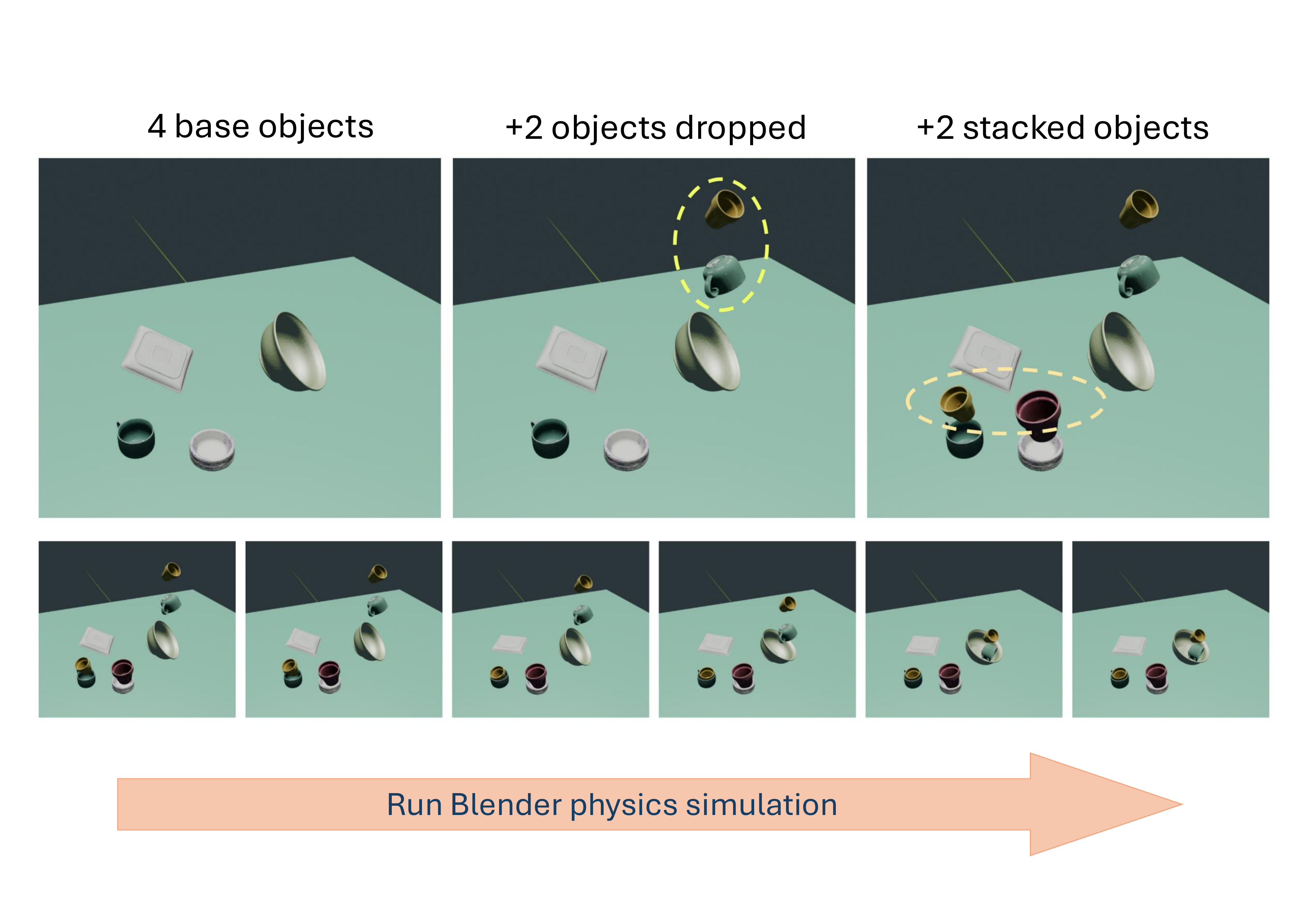}
    \caption{
    An example of controlled scene generation for the "hard" category. 
    }
    \label{fig:mk_MK-synthetic-creation-fig}
\end{figure*}

\subsection{Controlled scene generation}

We generate synthetic scenes following the same difficulty levels as in the Messy\-Kitchens-Real dataset. 
For easy scenes, we randomly place or drop four objects onto a flat plane. To increase realism and diversity, we enforce a balanced distribution of object orientations: approximately 50\% of base objects are placed upright, reflecting common kitchen arrangements, while the remaining objects are placed with arbitrary orientations. 

For medium scenes, we first sample four base objects and randomly perturb their $(x,y)$ positions around the scene center while placing them appropriately along the $z$-axis. The extent of perturbation in X and Y axes is also a parameter for us to control how tight the medium scenes would be.

For hard scenes, We then select two additional objects and situate them above the existing support objects so that when gravity acts on it during simulation, we get stacks of objects (number of stacked set of objects and number of objects in each stack is also controllable). Stacking feasibility is determined using a precomputed dictionary containing upright top-face bounding box areas and object volumes. Objects are only stacked if their volume and support surface area are compatible. Furthermore, we maintain a list of non-stackable objects (e.g., pitchers, kettles) that are never placed on top of other objects. For hard scenes, we extend this procedure to allow multiple stacked and nested configurations, carefully controlling stacking and compatibility to produce realistic, cluttered arrangements. Figure \ref{fig:mk_MK-synthetic-creation-fig} shows how a typical "hard" scene is created.

\begin{figure*}[t]
    \centering
    \includegraphics[width=\textwidth]{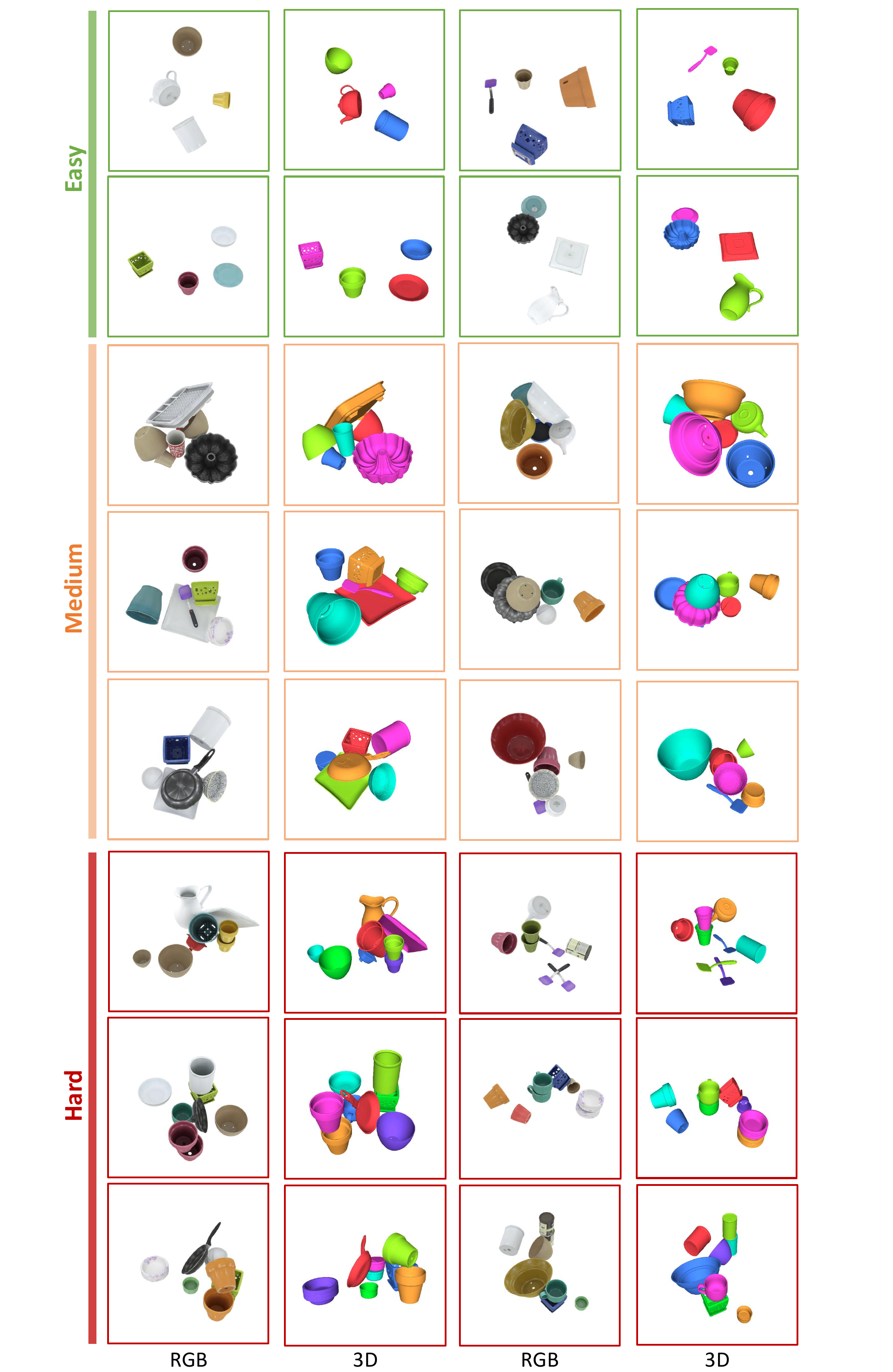}
    \caption{
    A few samples from all the three difficulty levels of our MessyKitchens-Synthetic dataset. We show the rendered RGB image and the ground-truth 3D of the scene obtained from our Blender simulation.
    }
    \label{fig:mk_dataset_samples_synth}
\end{figure*}

In Figure \ref{fig:mk_dataset_samples_synth} we show a few samples of our MessyKitchens-Synthetic dataset from all the three difficulty level categories -- Easy, Medium and Hard. %

\section{MessyKitchens dataset summary}
\label{sec:app_MKDataset_summary}

We compare our dataset statistics with other benchmarks in Table \ref{tab:app_MKDataset_summary}.  In addition to being highly accurate in terms of registration accuracy that naturally yields accurate contacts, MessyKitchens includes a high number of synthetic scenes and kitchenware objects, benefiting specific domain knowledge.  

\begin{table}[h]
\caption{MessyKitchens dataset overview.}
\label{tab:app_MKDataset_summary}
\centering
\resizebox{\linewidth}{!}{
\begin{tabular}{lcccccc}
\hline
\textbf{Datasets} & \textbf{registration accuracy} & \textbf{\#real scenes} & \textbf{\#object instances} & \textbf{\#kitchenware objects} & \textbf{\#synthetic scenes} \\
\hline
GraspNet-1B [18]     & 7.69 mm       & 190          & 88  & 26    & 0 \\
HouseCat6D [31]      & 1.73 mm       & 41           & 192 & $\sim$110 & 0 \\
GraspClutter6D [3]   & 3.22 mm       & \textbf{1000} & 200 & $\sim$63  & 0 \\
YCB-Video [55]       & 3.95 mm       & 92 (videos)  & 21  & 13    & 0 \\
T-LESS [25]          & 4.28 mm       & 20           & 30  & 0     & 0 \\
LINEMOD [24]         & 5.89 mm       & 15 (video)   & 15  & 0     & 0 \\
MP6D [10]            & 3.54 mm       & 11           & 20  & 0     & 50 \\
Omni6DPose [61]      & Not specified & 363 (videos) & \textbf{581} & 59 & 50 \\
\hline
\textbf{MessyKitchens} & \textbf{1.62 mm} & 100 & 130 & \textbf{130} & \textbf{1800} \\
\hline
\end{tabular}
}

\end{table}

\end{document}